\documentclass[conference]{IEEEtran}
\IEEEoverridecommandlockouts
\usepackage{cite}
\usepackage{amsmath,amssymb,amsfonts}
\usepackage{algorithmic}
\usepackage{graphicx}
\usepackage{textcomp}
\usepackage[colorlinks=true,linkcolor=blue, urlcolor=blue]{hyperref}%
\usepackage{xcolor}
\def\BibTeX{{\rm B\kern-.05em{\sc i\kern-.025em b}\kern-.08em
    T\kern-.1667em\lower.7ex\hbox{E}\kern-.125emX}}
\begin{document}

\title{MOFit: A Framework to reduce Obesity using Machine learning and IoT\\
{\footnotesize \textsuperscript{*}Note: This paper is accepted  in the 44th MIPRO 2021 convention. The final version of this paper will appear in the conference proceedings.}
}

\author{\IEEEauthorblockN{ Satvik Garg}
\IEEEauthorblockA{\textit{Department of Computer Science} \\
\textit{Jaypee University of Information Technology}\\
Solan, India \\
satvikgarg27@gmail.com}
\and
\IEEEauthorblockN{Pradyumn Pundir}
\IEEEauthorblockA{\textit{Department of Computer Science} \\
\textit{Jaypee University of Information Technology}\\
Solan, India  \\
pundirpradyumn25@gmail.com}
}

\maketitle

\begin{abstract}
From the past few years, due to advancements in technologies, the sedentary living style in urban areas is at its peak. This results in individuals getting a victim of obesity at an early age. There are various health impacts of obesity like Diabetes, Heart disease, Blood pressure problems, and many more. Machine learning from the past few years is showing its implications in all expertise like forecasting, healthcare, medical imaging, sentiment analysis, etc. In this work, we aim to provide a framework that uses machine learning algorithms namely, Random Forest, Decision Tree, XGBoost, Extra Trees, and KNN to train models that would help predict obesity levels (Classification), Bodyweight, and fat percentage levels (Regression) using various parameters. We also applied and compared various hyperparameter optimization (HPO) algorithms such as Genetic algorithm, Random Search, Grid Search, Optuna to further improve the accuracy of the models. The website framework contains various other features like making customizable Diet plans, workout plans, and a dashboard to track the progress. The framework is built using the Python Flask. Furthermore, a weighing scale using the Internet of Things (IoT) is also integrated into the framework to track calories and macronutrients from food intake.
\end{abstract}


\section{Introduction}
According to an article published in GBD 2017 Obesity Collaborators [1], over 4 million people die each year due to excessive weight. From 1975 to 2016, the number of children and adults experiencing obesity had increased from 4\% to 18\%. The problem of overweight and obesity is not only a problem of developed countries, even low-income and developing countries are overtaking the total number which is 30\% excess than the total cases in developed countries [2].

The effects of being overweight have different measurements, from psychological health issues to trivial medical conditions. Nonetheless, there is no real link between emotional health and obesity. However, due to isolation, there is a greater risk of low self-confidence, mood, and motivational issues, eating disorders, relational communication issues, and all of which directly or implicitly affect personal satisfaction [3]. Physical medical conditions include type 2 diabetes, high blood pressure, heart and kidney diseases, asthma, back pain, and more. This prompts a total of more than 3,000,000 passings yearly [4].

Body mass index (BMI) is a person's weight in kilograms divided by the square of height in meters. A high BMI over 25 can be a marker of high body weight. Daily living factors such as eating pattern, level of exercise, total water intake, alcohol intake and unhealthy food intake, and so forth fully contribute in deciding an individual's body mass index [5]. Predicting obesity levels using these variables sets people up to think about these factors from top to bottom and helps link habits that encourage them to lose weight. BMI can be utilized to screen for weight classifications that may indicate medical conditions yet are not a good indicator of a person's body fatness. Studies have shown that BMI is quite outdated not to consider lean body mass to calculate overall fitness [6]. Two people with the same weight and height may look different based on lean body mass present. Generally, the higher the muscle mass, the higher a person's metabolism is resulting in lower body fat levels. Various factors such as gender, arm thickness, waist circumference are good characteristics to tell the body fat percentage of an individual.

Diet or proper food intake plays a dominant role in getting healthy. However, there is going around a lot of confusion in the fitness health industry regarding diet plans which are not sustainable in the long term [7]. Improper knowledge about calories coming from macro-nutrients and micro-nutrients in food leading individuals to get demotivated as they do not see any progress and they end up consuming more or less food according to their goals. So a proper framework website is required which predicts the BMI or obesity levels, body fat percentage and suggest the dietary requirements according to their goals. A smart IoT weighing scale system is designed and integrated into the framework that tells the calories generated from food. A dashboard has also been created which gives line plots to track the daily progress made by an individual in getting healthy. For the machine learning part, algorithms like Random Forest, KNN, XGBoost, Extra Trees, and Decision Tree are used to predict obesity, weight, and fat levels. The algorithms are then optimized using the Genetic algorithm, Random Search, Grid Search, Optuna. The website is built using the python flask framework and various components like HX711, load cell, NodeMCU are used to build a weighing scale.

This examination work is organized as follows: Section II contains the related works dependent on health frameworks. Segment III includes the methodology utilized for building the models and systems. The examination and assessment of predictions using different metrics are introduced in Section IV. Segment V contains the functionalities of the website framework. Presented an IoT weighing scale working in Section VI. Section VII concludes the paper.

\section{Related Works}

A good amount of work has been done on the data produced from sensors and trackers to identify human activity, with various applications such as tracking all body movements, counting calories consumed while working out, estimating the pulse, and applications in the clinical field [8]. However, in the field of overall welness and general health, work is not carried out at a high pace. Despite this, few studies and examinations focused on common health issues such as calorie estimation, IoT-based smart health systems, BMI estimations.

Raza Yunus et al. [9] developed a mobile-based dietary assessment framework application that can record real-time photographs of food and examine them for nutritional value that improves dietary intake. Deep learning techniques for image classification were performed on a dataset consisting of 100 classes with an average of 1000 images for each class. The Food-101 dataset was also combined to include subcontinental food. The results show that the completed model accuracy is 85 percent and is similarly productive on the Essential Food 101 dataset.

%There has been a rapid increase in dietary ailments during the last few decades, caused by unhealthy food routines. Mobile-based dietary assessment systems that can record real-time images of meals and analyze them for nutritional content can be very handy and improve dietary habits, and therefore, result in a healthy life. This paper proposes a novel system to automatically estimate food attributes such as ingredients and nutritional value by classifying the input image of food. Our method employs different deep learning models for accurate food identification. In addition to image analysis, attributes and ingredients are estimated by extracting semantically related words from a huge corpus of text, collected over the Internet. We performed experiments with a dataset comprising 100 classes, averaging 1000 images for each class to acquire a top 1 classification rate of up to 85 percent. An extension of a benchmark dataset Food-101 is also created to include sub-continental foods. Results show that our proposed system is equally efficient on the basic Food- 101 datasets and its extension for sub-continental foods. The proposed system is implemented as a mobile app that has its application in the healthcare secto
Considering the effect of BMI on people in everyday activities, Enes Kocabey et al. [10] constructed a model name FACE2BMI which foresee the BMI levels dependent on the face of a person. The visual-BMI dataset was used. It contains an aggregate of 16483 pictures which further decreased to 4206 pictures after the manual selection. The Pre-trained CNN models VGG-Net and VGG-Face are used for the feature extraction part and the latter incorporates a support vector regression model for BMI prediction. The outcomes show that VGG-Face accomplished an overall accuracy of 65\% contrasted with 47\% utilizing VGG-Net. 

%A person's weight status can have profound implications on their life, ranging from mental health to longevity, to financial income. At the societal level, "fat-shaming" and other forms of "sizeism" are a growing concern while increasing obesity rates are linked to ever-rising healthcare costs. For these reasons, researchers from a variety of backgrounds are interested in studying obesity from all angles. To obtain data, traditionally, a person would have to accurately self-report their body-mass index (BMI) or would have to see a doctor to have it measured. In this paper, we show how computer vision can be used to infer a person's BMI from social media images. We hope that our tool, which we release, helps to advance the study of social aspects related to body weight.

Sebastian Baumbach et al. [11] analyzed compound movements like Deadlifts and Bench press typically directed with athletic gear in the fitness center. The author gathered sensor information of 23 members for these exercises using smartwatches and smartphones. Conventional machine learning and deep learning algorithms were used for training. SVM and Naive Bayes performs best with an exactness of 80\%, although deep learning models overachieved them with a precision of 92\%. 

Mehak Gupta et al. [12] proposed a model to predict future weight levels dependent on the clinical history of childhood. This is useful in the early identification of youth at high risk of developing childhood obesity which may help link earlier and more viable mediation to prevent adulthood obesity. LSTM were utilized on unaugmented Electronic Health Record (EHR) dataset. Set embedding and attention layers were also used to add interpretability at both the features and timestamp levels.

%We	adopt	a	general	LSTM	(long	short-term	memory)	network	architecture for	our	model	for	training	over	dynamic	(sequential)	and	static	(demographic)	EHR	data.	We	have	additionally included a	set	embedding	and	 attention	layers to	compute	the	feature	ranking	of	each	timestamp	and	attention scores	of	each	hidden	layer	 corresponding	 to	 each	 input	 timestamp.	 These	 feature	 ranking	 and	 attention	 scores added interpretability	at	both	the	features	and	the	timestamp-level.

%The objective of this project is to design a smart fitness tracker system that will record users' indoor fitness routines by minimizing the hassle of manually counting the sets and repetitions, especially of weight exercises. An attempt has been made to distinguish between each member of the gym using Radio Frequency Identification (RFID) scanner and tag. This aids in separately feeding and updating users' respective data on the database which will further contribute to the Fitness Tracker App. This project can help solve problems of the traditional way of mentally logging the data which tends to be inaccurate. We analyze the recorded data to give suggestions about users' progress which helps in setting a goal for the future. This project comes under the category of Green IT since it uses cloud services. Furthermore, many such trackers for various equipment can be evolved from this project

\section{Methodology}
The datasets used in this research are 'Estimation of obesity levels based on eating habits and physical condition dataset' taken from UCI ML repository [13]  and body fat percentage dataset [14] published in Journal of Statistics Education. The obesity dataset consists of 2111 rows and 17 columns whereas the body fat percentage dataset has 252 rows and 19 columns. The description of both datasets are given in Fig. 1 and 2 respectively. They are divided using 80:20 train test split to build three machine learning models namely, Obesity level prediction (Classification), Weight prediction (Regression), and Body fat percentage prediction (Regression) as seen in Table I.
\begin{figure}[htbp]
\centerline{\includegraphics[width=8cm, height=3.5cm]{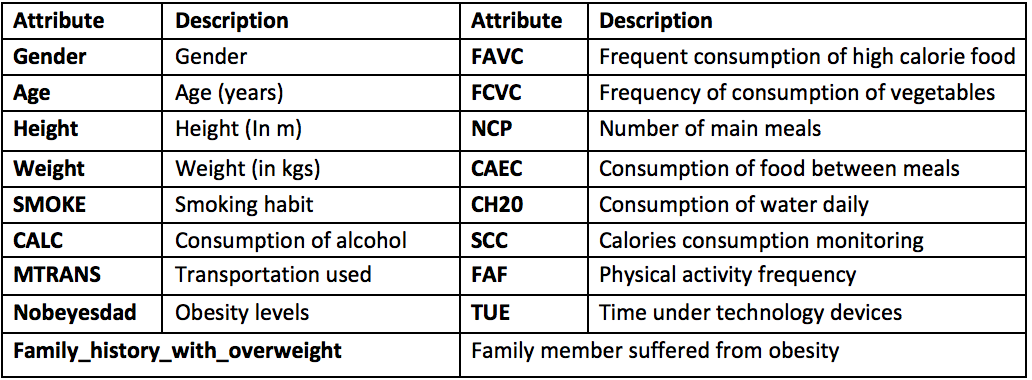}}
\caption{Description of obesity level dataset}
\end{figure}

\begin{figure}[htbp]
\centerline{\includegraphics[width=8cm, height=3.5cm]{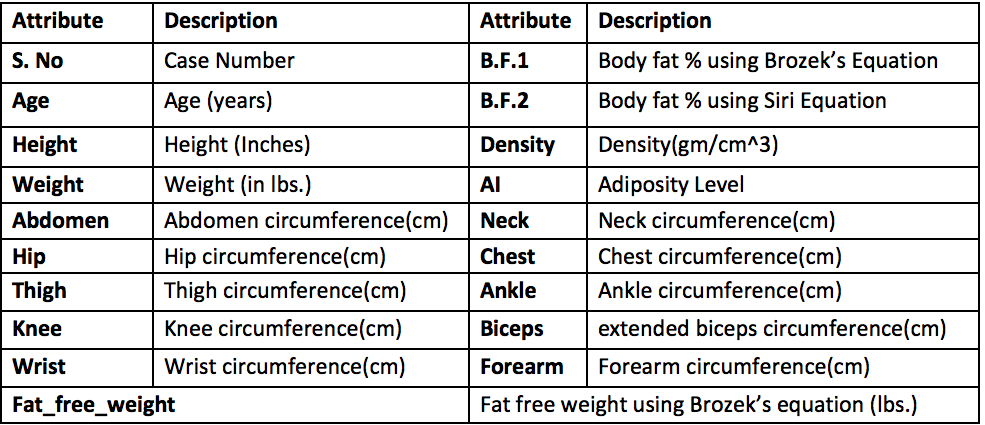}}
\caption{Description of Body fat percentage dataset}
\end{figure}

\begin{table}[htbp]
\caption{Distribution of Dataset}
\begin{center}
\begin{tabular}{|p{2.5cm}|p{1.6cm}|p{1.5cm}|p{1.5cm}|}
\hline
\textbf{Task} & \textbf{Dataset used}&\textbf{Train (80\%)}& \textbf{Test (20\%)}\\
\hline
Prediction of Obesity level& Obesity Level & 1688& 423\\
\hline
Prediction of Weight (kg)& Obesity Level&1688& 423\\
\hline
Prediction of Body Fat percentage & Body Fat &201&51\\
\hline
\end{tabular}
\end{center}
\end{table}
The flowchart of the proposed pipeline is given in Fig. 3. It contains a total of four parts namely, data preparation, training (Algorithms), hyper-optimization, and testing part (metrics).

\begin{figure}[htbp]
\centerline{\includegraphics[width=7cm, height=7cm]{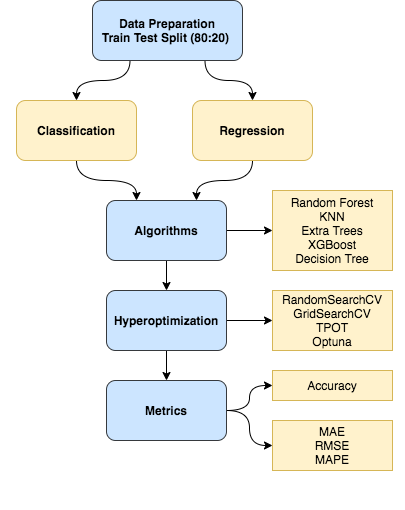}}
\caption{Flowchart of the methodology adopted}
\end{figure}

\subsection{Data Preparation}

The initial stage of every machine leaning problem is the data preparation part. Without appropriate information handling methods, the model will restore garbage values and will not be effective. We employed common data preprocessing strategies such as; Inspect null values, duplicate columns and rows, eliminate null values, and text from tuples. The obesity level dataset contains categorical features converted to numerical features using label and ordinal encoding methods. The ordinal encoding method is used to preserve the order of categorical data whereas the label encoding method is used when there is no order in data. We manually checked the order using the cat plot imported from the seaborn library [15]. As demonstrated in Fig. 4, the consumption of alcohol (CALC) feature versus the BMI cat plot, there is no order present. This results in the one-hot encoding of the CALC feature. 
\begin{figure}[htbp]
\centerline{\includegraphics[width=8.5cm, height=3.5cm]{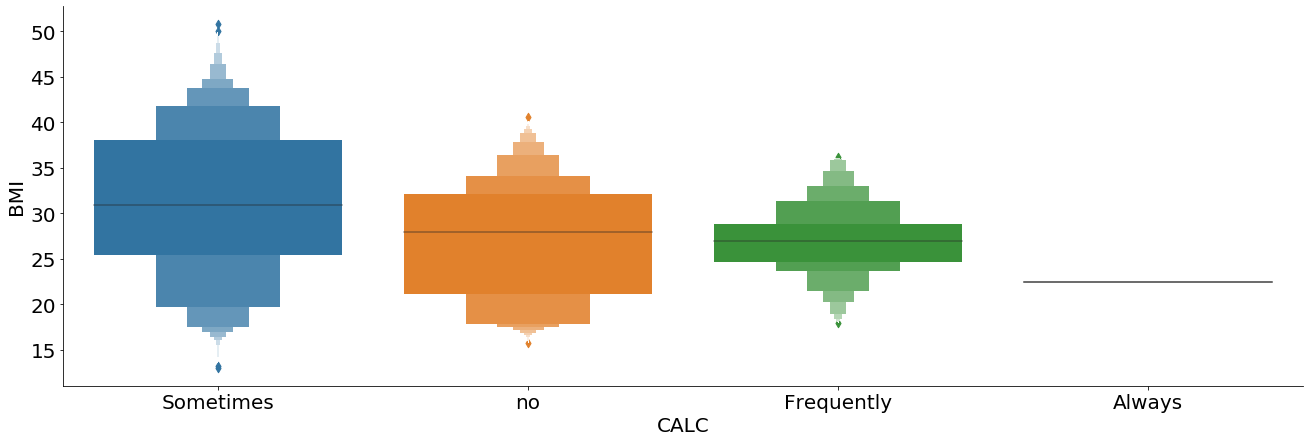}}
\caption{Cat Plot of CALC  w.r.t to BMI}
\end{figure}

For user usability, some features were removed before the machine learning algorithm was trained. For the obesity level prediction model, the target variable is NObeyesdad (obesity level). The weight and height features are reduced as we only require users to input their daily habits to obtain an estimated level of obesity. For the body weight expectation model, the objective variable is the weight and no other attributes are left for training. Also added the height parameter as it is imperative to get the expected predicted value near the actual value. The user is first able to estimate the level of obesity using daily habits and later inputs an estimated estimate of height and obesity to obtain the expected value of body weight. In body fat percentage model, the BF1 and BF2 features are combined to obtain an average target variable feature BF. Similarly, BF1 and BF2 were dropped, and later features such as AI, Density, and Fat Free Weight were also dropped for user convenience.

%ek chota sa flowchart and then explain. Similarly, for features like bidsv s label encoding nominal encoding ,some visualizations male vs female ka visualization according to BMI ,The first step is to divide the dataset into train and test split (80:20).
\subsection{Algorithms}
To be specific, for both classification and regression, a total of five algorithms, Random Forest, KNN, XGBoost, Extra Tree and Decision Tree are employed to build the prediction models. Deep learning models such as neural networks, in general, outperform any other machine learning algorithm. But with respect to small to medium datasets, tree-based algorithms are seen as the top tier at the present time [16]. 
\subsubsection{Decision Tree}
A decision tree [17] is a tree where each node addresses a feature i.e. attribute, each link or branch corresponds to a decision and each leaf addresses an outcome which is categorical or real value based on the machine learning problem. 
The goal of using a decision tree is to create an explanatory model that can be used to estimate the class or values of the target variable by taking the required decision parameters interpreted from the previously generated information.
In Decision Trees, the initial step in predicting the target class for a record was to check the estimation of the original attribute with the record attribute. Based on the correlation, the branch is followed and jumped to the following node compared to that value.
\subsubsection{Random Forest}
Random Forest [18] is based on the concept that it includes various decision trees on different subsets of the given dataset and took the mean to improve the prescient exactness of that dataset. Instead of depending on one decision tree, it takes the forecast from each tree, and depending on the dominant part votes of expectations it predicts the final output. Random forest relies on the idea of ensemble learning, which is a cycle of merging numerous classifiers to solve difficult problems and to improve the performance of the model. 
\subsubsection{KNN}
The K-nearest neighbor algorithm [19] analyzes the similarity between features to estimate the projections of new data-points, which means that it depends on how closely it coordinates the points in the preparation set. It is likewise called lazy learner computation because it does not rapidly benefit from the preparation set. 
\subsubsection{XGBoost}
Extraordinary gradient boosting or XGBoost [20] for short is one of the notable ensemble-based gradient boosting techniques having upgraded execution and speed in tree-based machine learning algorithms. The idea of gradient boosting algorithms is to optimize its loss function, unlike other boosting algorithms where the aim is to expand the weights of misclassified branches. XGBoost is a high-level execution of gradient boosting along with some regularization factors that help smooth the weights to avoid over-fitting.
\subsubsection{Extra Trees}
Extra Trees or extremely randomized trees [21] are based on an ensemble machine learning algorithm. This algorithm works by making an enormous number of unpruned decision trees using the training data. For regression-based problems, the predictions generated by decision trees are averaged or in case of classification, majority voting is used. Unlike a random forest that builds each decision tree from the bootstrap sample of the training dataset, the additional tree fits each decision tree onto the overall training dataset. The extra tree algorithm chooses the random split point which prefers greedy algorithm, and it can arbitrarily samples the features at each split point.

\subsection{Hyperoptimization}
Machine learning algorithms have hyperparameters that must be set to transform and improve the accuracy of the model. The overall effect of hyperparameters on a model is known, yet it is difficult to figure out on how to best set up a mix of hyperparameters and communication hyperparameters for a given dataset. A better method is to objectively scan various properties for model hyperparameters and choose a subset that results in a model that meets the best performance on a given dataset [22]. This is called Hyperparameter advancement or Hyperparameter Tuning and can be accessed in the scikit-learn Python ML library [23]. The hyperoptimization system involves marking a search space [24]. It can be thought of mathematically as an n-dimensional volume, where each hyperparameter addresses an optional dimension and the range of the dimensions are the values hyperparameters can take, for example, categorical or real-valued. A point that lies in the search space is a vector with a particular subset representing each hyperparameter. The aim of optimization is to find a vector that results in the best performance of the model after training with high accuracy or least error.

\subsubsection{Random Search}
Random search [25] characterizes a search space as a finite region of hyperparameter values and randomly selects points in that space. It establishes a set of hyperparameter values and chooses arbitrary combinations that completely nullifies the total choice of combinations to produce the model and score.
This allows one to implicitly control the number of parameters attempted. It is conceivable that RandomSearch will not yield the exact same result as GridSearch, but generally it picks the best result in a small fraction of the time. 

\subsubsection{Grid Search}
The customary technique for hyperparameter optimization is a grid search [26], which essentially performs an aggregate search on a given subset of the parameter space of the training algorithm. Since the boundary space can include spaces with real or infinite properties for some range, it is conceivable that we need to set a boundary to implement a grid search. Grid search experiences higher dimensional spaces, although they can undoubtedly be parallelised because the hyperparameter values with which the algorithm works are usually autonomous from each other.
\subsubsection{Genetic Algorithm}
The genetic algorithm [27] is an evolutionary search algorithm used to take care of growth and performance by sequentially selecting, incorporating, and changing boundaries using processes that occur after biological evolution. 
Evolutionary algorithms mimic the cycle of natural selection which refers to species that can adjust to changes in the environment and tolerate and emulate future generations. 
Each generation comprises a population of individuals and each individual constitutes a point in the search space and represents viable solution. Every individual is addressed as a string of character/number/bits and is practically equivalent to the chromosome. The genetic calculation starts with a randomly produced populace of chromosomes. At that point, it undergoes a mechanism of selection and recombination, which is based on chromosome score of fitness. Parent hereditary materials are recombined to create child chromosomes creating the next generation. This cycle is repeated up until a stopping criterion is satisfied.

\subsubsection{Optuna}
Optuna [28] is a framework built for automating the process of optimization.  It consequently finds ideal hyperparameter values by utilizing various samplers, for example, random, grid search, bayesian, and genetic calculations. It underscores a definitive, characterize by-run way to deal with client API. Because of the defined by-run API, the code content composed with Optuna holds modularity, and the Optuna clients could effectively form search spaces for the parameters. Optuna depicts each interaction as a \textit{study} which is an improvement that depends on the objective function, and \textit{trial} which is a single assessment or execution of the objective function.

\subsection{Metrics}
For measuring the exactness of obesity level, the accuracy metric is used given by:
\begin{equation}
Accuracy = \frac{T_{p}+T_{n}}{T_{p}+T_{n} + F_{p}+F_{n}}
\end{equation}
where $T_{p}$, $T_{n}$, $F_{p}$, $F_{n}$ represents True-positive, True-negative, False-positive and False-negative respectively. Three metrics namely, root mean squared error (RMSE), mean absolute error (MAE), mean absolute percentage error (MAPE) are employed for analyzing the predictions of Bodyweight and fat percentage levels.
\begin{equation}
    RMSE = \sqrt{\frac{1}{n}\sum_{i=1}^{n}{(x_{i}-y_{i})^2}}
\end{equation}
\begin{equation}
    MAE = (\frac{1}{n})\sum_{i=1}^{n}\left | (x_{i} - y_{i} \right)|
\end{equation}
\begin{equation}
    MAPE = (\frac{100}{n})\sum_{i=0}^n \frac{\left | x_{i} - y_{i} \right |}{\left | x_{i}  \right |}
\end{equation}
where, $x_{i}$ : actual values, $y_{i}$ : predicted values, $n$ : length of test data.

\section{Results}
In this research, three prediction tasks are performed utilizing five algorithms specifically, Random Forest, KNN, XGBoost, Decision Tree, and Extra Trees. These models are then optimized using four HPO techniques to be specific, Random Search, Grid Search, Genetic algorithm, Optuna. The Random Search and Grid search are performed using RandomSearchCV and GridSearchCV functions provided by scikit library [24]. The genetic algorithm is implemented by the Tree-based Pipeline Optimization Tool (TPOT) [29]. The TPOT is constructed using a genetic search algorithm and is an automated machine learning tool. For Optuna, an objective function is created to return accuracy in which hyper-parameters are proposed using the trial object and final execution is finished with the study object.

Table II shows the accuracy of the classification of obesity levels on the test dataset. It suggests the random forest model using grid search and Tpot classifier outperformed all other models with 86\% accuracy. The algorithm with the lowest performance is the decision tree which range between 73-75\% accuracy. The Extra tree classifier gave no indication of progress in accuracy and it just stuck with 85\% accuracy. All HPO models except Random Search are producing good results.

\begin{table}[htbp]
\caption{Accuracy of predicted Obesity level}
\begin{center}
\begin{tabular}{|c|c|c|c|c|}
\hline
\textbf{Algorithm} & \textbf{RSearch}& \textbf{Gsearch}& \textbf{Tpot}& \textbf{Optuna}\\
\hline
RandomForest& 0.84& 0.86&0.86 &0.85 \\
\hline
DecisionTree& 0.73& 0.76&0.74 &0.73\\
\hline
ExtraTrees& 0.85& 0.85&0.85 &0.85 \\
\hline
KNN& 0.81& 0.82&0.82 &0.82\\
\hline
XGBoost& 0.84& 0.84&0.84 &0.85\\
\hline
\end{tabular}
\end{center}
\end{table}

The body weight predictions dissected by three metrics namely RMSE, MAE, MAPE are given in Table III. It is observed that there are some instances where increase or decrease in RMSE shows opposite effect on MAE and MAPE. For example, in the decision tree algorithm using Tpot, the RMSE and MAE values are 3.50 and 2.23, respectively. Applying Optuna, the RMSE value expanded to 3.88, while the MAE decreased to 2.13, which recommends that during training the model concentrates more on correcting small values and neglects to correct larger errors which was further penalized by RMSE. The results shows that the KNN algorithm is giving the highest error whereas the Extra Tree Resistor postprocessed by Optuna has been chosen for the framework application because of the minimum MAE and MAPE values.

\begin{table}[htbp]
\caption{Metrics of predicted Body Weight}
\begin{center}
\begin{tabular}{|c|c|c|c|c|c|}
\hline
\textbf{Model} & \textbf{Metric}& \textbf{RSearch}& \textbf{GSearch}& \textbf{Tpot}& \textbf{Optuna}\\
\hline
& RMSE& 2.81&2.80 &2.86&2.85 \\
RandomForest& MAE& 1.65&1.64 &1.71&1.67\\
& MAPE& 0.022&0.022 &0.023&0.022\\
\hline
& RMSE& 3.80&3.75 &3.50&3.88\\
DecisionTree& MAE& 1.99&2.29 &2.23&2.13\\
& MAPE& 0.026&0.030 &0.029&0.027\\
\hline
& RMSE& 2.89&2.86 &2.89&2.82 \\
ExtraTree& MAE& 1.62&1.62 &1.62&1.57\\
& MAPE& 0.022&0.022 &0.022&0.021\\
\hline
& RMSE& 5.38&5.30 &5.02&5.44\\
KNN& MAE& 3.16&2.98 &3.13&3.10\\
& MAPE& 0.044&0.041 &0.042&0.043\\
\hline
& RMSE& 3.02&2.96 &2.95&2.87\\
XGBoost& MAE& 1.97&1.84 &1.87&1.80\\
& MAPE& 0.026&0.024 &0.025&0.023\\
\hline
\end{tabular}
\end{center}
\end{table}

For prediction of body fat percentage, as shown in Table IV, the XGBoost algorithm outperforms all other algorithms that demonstrate its merit in small datasets. As far as HPO model performance is concerned, Optuna outperforms other HPO techniques on all algorithms. XGBoost with Optuna has been chosen for the deployment of the Body Fat Percentage Model as it is giving minimum errors with respect to all metrics. KNN again failed to meet the expectations in contrast to other algorithms.

\begin{table}[htbp]
\caption{Metrics of predicted Body Fat Percentage}
\begin{center}
\begin{tabular}{|c|c|c|c|c|c|}
\hline
\textbf{Model} & \textbf{Metric}& \textbf{RSearch}& \textbf{GSearch}& \textbf{Tpot}& \textbf{Optuna}\\
\hline
& RMSE& 3.81&3.82 &3.85&3.80 \\
RandomForest& MAE& 3.13&3.16 &3.17&3.11\\
& MAPE& 0.237&0.235 &0.234&0.233\\
\hline
& RMSE& 4.48&4.20 &4.54&3.90\\
DecisionTree& MAE& 3.53&3.36 &3.62&3.05\\
& MAPE& 0.259&0.246 &0.263&0.234\\
\hline
& RMSE& 3.77&3.78 &3.78&3.74 \\
ExtraTree& MAE& 3.11&3.11 &3.12&3.10\\
& MAPE& 0.232&0.230 &0.232&0.228\\
\hline
& RMSE& 4.45&4.45 &4.40&4.45\\
KNN& MAE& 3.72&3.72 &3.52&3.72\\
& MAPE& 0.276&0.276 &0.252&0.276\\
\hline
& RMSE& 3.61&3.74 &3.67&3.47\\
XGBoost& MAE& 2.92&3.14&2.99&2.80\\
& MAPE& 0.213&0.233 &0.215&0.208\\
\hline
\end{tabular}
\end{center}
\end{table}

To summarize, Grid Search goes for every conceivable combination of search space provided which makes it brute force and slow. It is not recommended to train a grid search if the search space is large. Random search trains a little faster but it does not ensure the best results. Tpot is a good option if the search space is large but it takes a lot of time if trained on large number of generations and population. To the best of our knowledge, Optuna is the suggested HPO model as it requires less time and gives top results. However, choosing the right parameter for an objective function is quite difficult. In any case, with respect to general performance, Optuna outpaces any remaining HPO models in terms of performance and results. 
In addition, different types of plots such as optimization history plots are provided by the optuna to analyze the hyperparameters as shown in Figure 5.

\begin{figure}[htbp]
\centerline{\includegraphics[width=9cm, height=5cm]{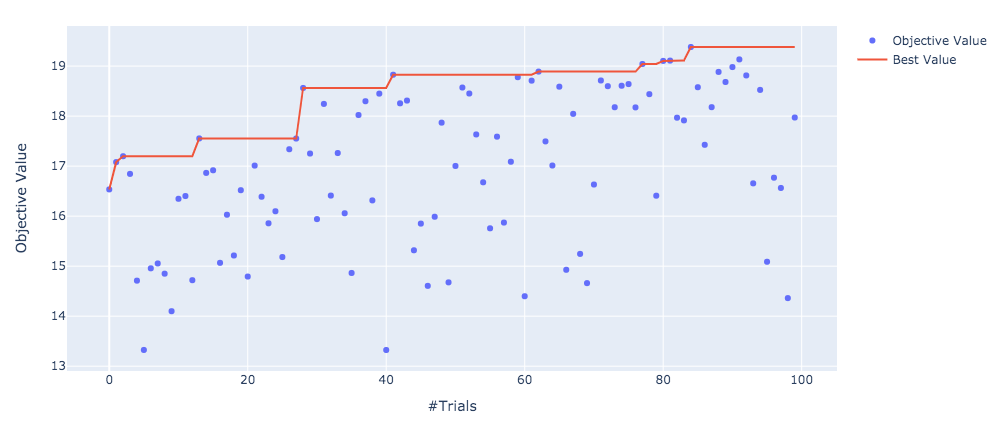}}
\caption{Optimization History Plot}
\end{figure}
%\begin{figure}[htbp]
%\centerline{\includegraphics[width=9cm, %height=4.5cm]{contour_mofit.png}}
%\caption{Contour Plot}
%\end{figure}

\section{MOFit Framework}
A Python Flask [30] is employed to deploy the machine learning models in a website framework. As we can see in Figure 6, the user first inputs the parameters on the GUI built using HTML, CSS to be sent as a query to the Flask framework. Flask will then use the trained model to predict values or labels on the input parameters and output the predicted results on the website page.
\begin{figure}[htbp]
\centerline{\includegraphics[width=8cm, height=2cm]{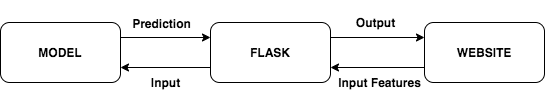}}
\caption{Model Deployment using Flask}
\end{figure}

To be specific a set of three algorithms were chosen, Random Forest, Xtra Tree, XGBoost for obesity, bodyweight and body fat prediction separately. Figure 7 shows the structure for the purpose of estimating obesity levels and body weight. The client first enters or selects an attribute such as age, gender, caloric probe, water consumption, etc. After clicking the submit button, the model predicted the obesity level for example, in this case, the obesity level ranges from BMI 30-35. Until then, the client needs to enter the height attribute in meters to effectively predict its weight which is 101kg as in Figure 7.
\begin{figure}[htbp]
\centerline{\includegraphics[width=8cm, height=7cm]{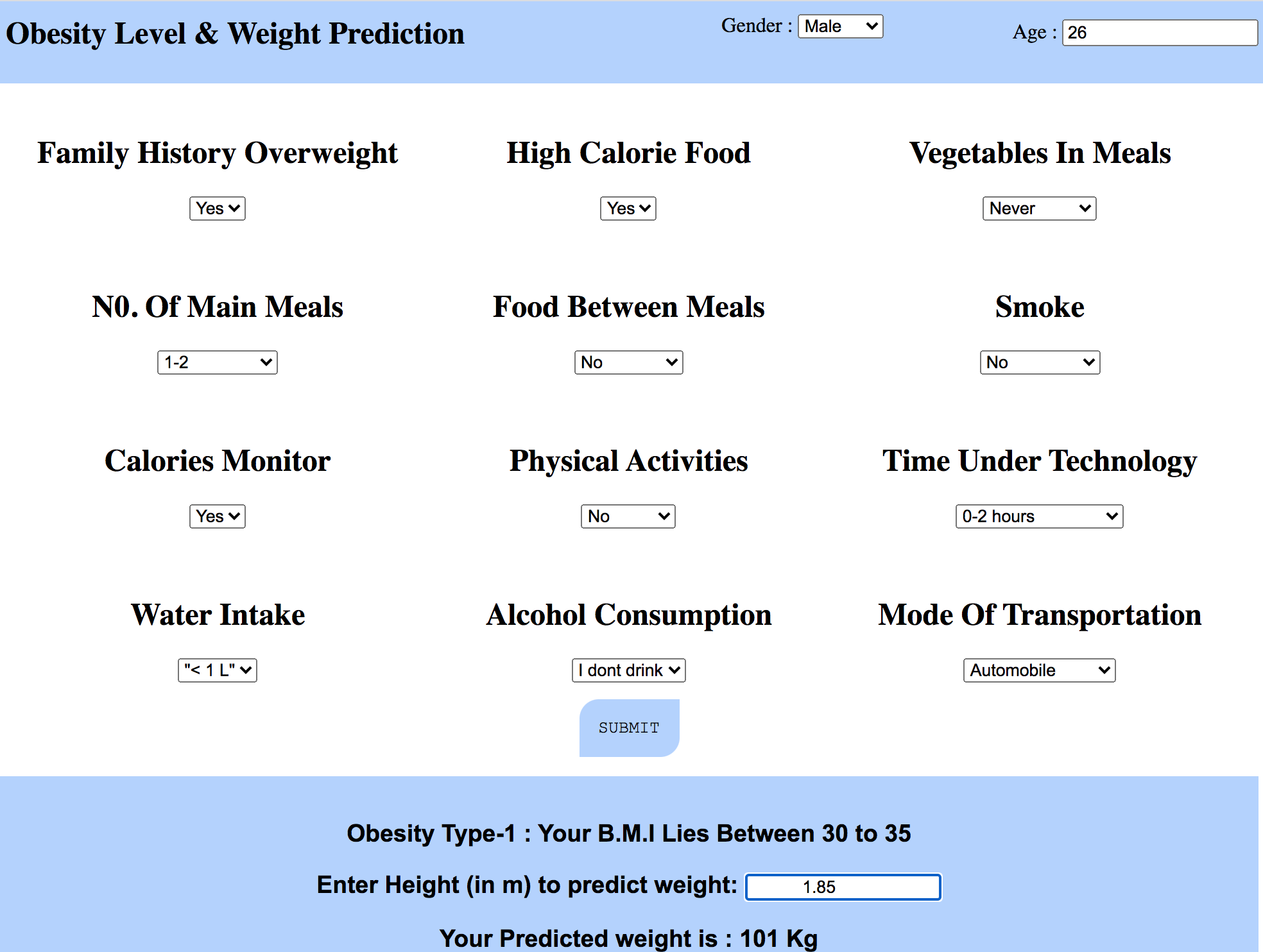}}
\caption{Structure of Deployed ML model to predict Obesity and Body Weight}
\end{figure}
%\begin{figure}[htbp]
%\centerline{\includegraphics[width=8cm, height=4.2cm]{bidyfat.png}}
%\caption{Optimization History Plot}
%\end{figure}

Another helpful component of the MOFit system is the dashboard where the customer can track its development every day, week to week and month to month. Line plots using the Canvas Js library [31] were used to plot the line plots of the actual weights versus the objective weights. This will help clients to realize how far behind their target weight they are. An example can be analyzed in Figure 8 where the customer has made steady progress on a weekly basis. The blue line shows the target progress weight while the red line shows the actual progress. The client began with 100Kg of weight, the ultimate objective load following a month is 80kg however the user covered till 86 kg.
\begin{figure}[htbp]
\centerline{\includegraphics[width=8cm, height=7cm]{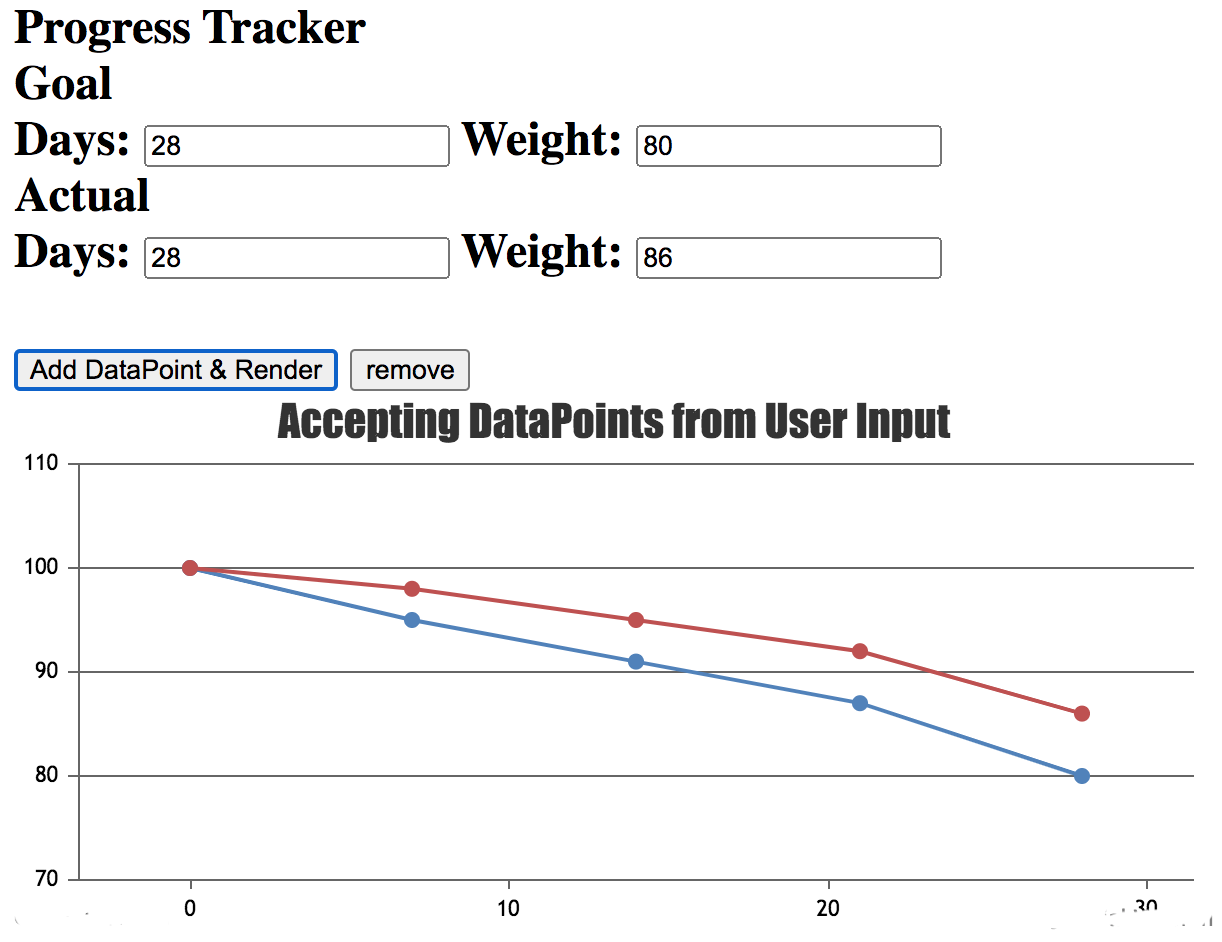}}
\caption{Progress Tracker}
\end{figure}

A customizable eating routine is additionally provided where the user can make better food choices and track calories. Having a balance in macronutrients i.e. proteins, carbohydrates and fats is fundamental to fulfilling certain purposes of body composition. In addition to calories, estimates of macronutrients are added to meet the need. As found in Figure 9, various food decisions can be made effectively, and a PDF option is given to create a PDF record to save the decisions for future reference.

\begin{figure}[htbp]
\centerline{\includegraphics[width=8cm, height=3cm]{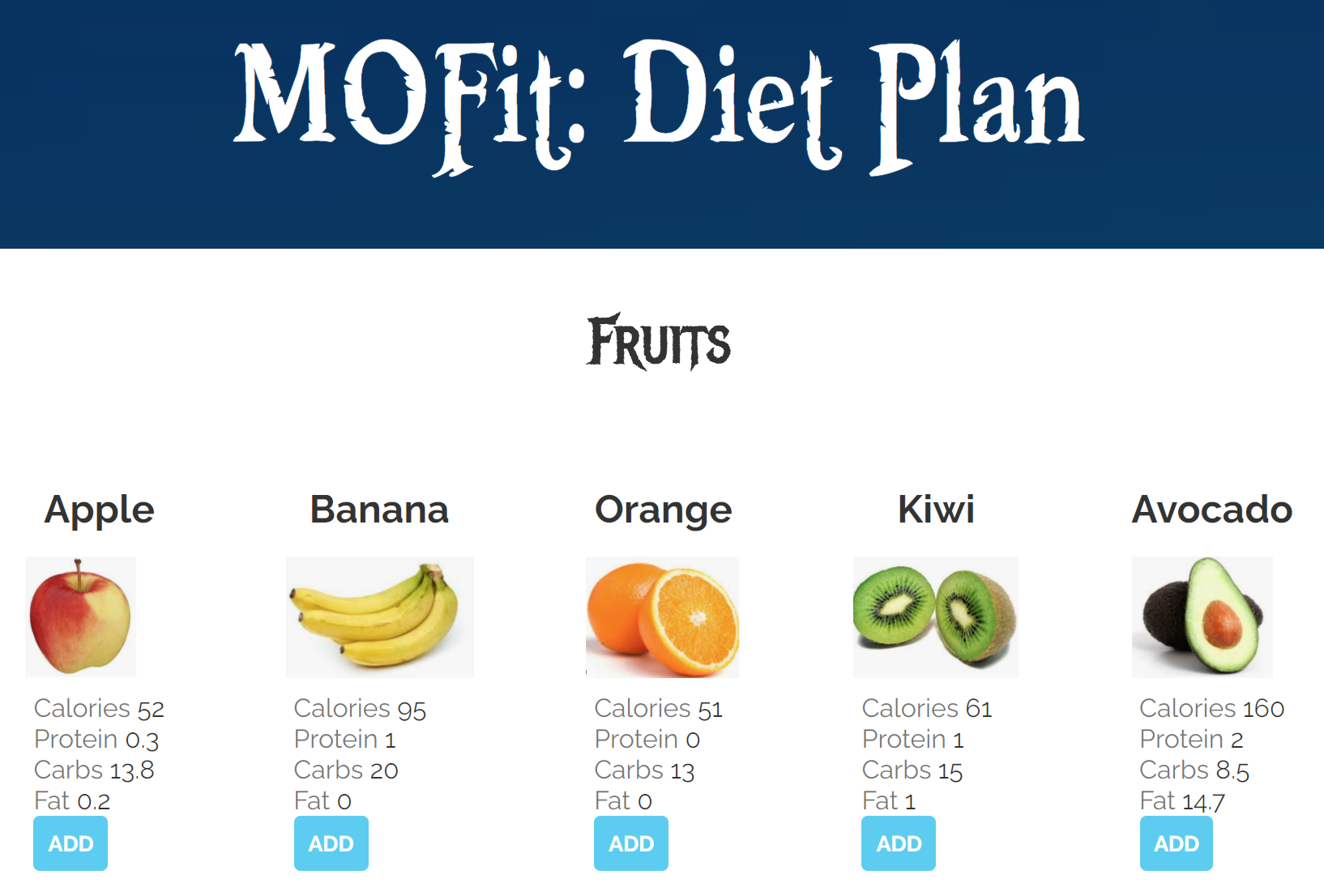}}
\centerline{\includegraphics[width=8cm, height=3cm]{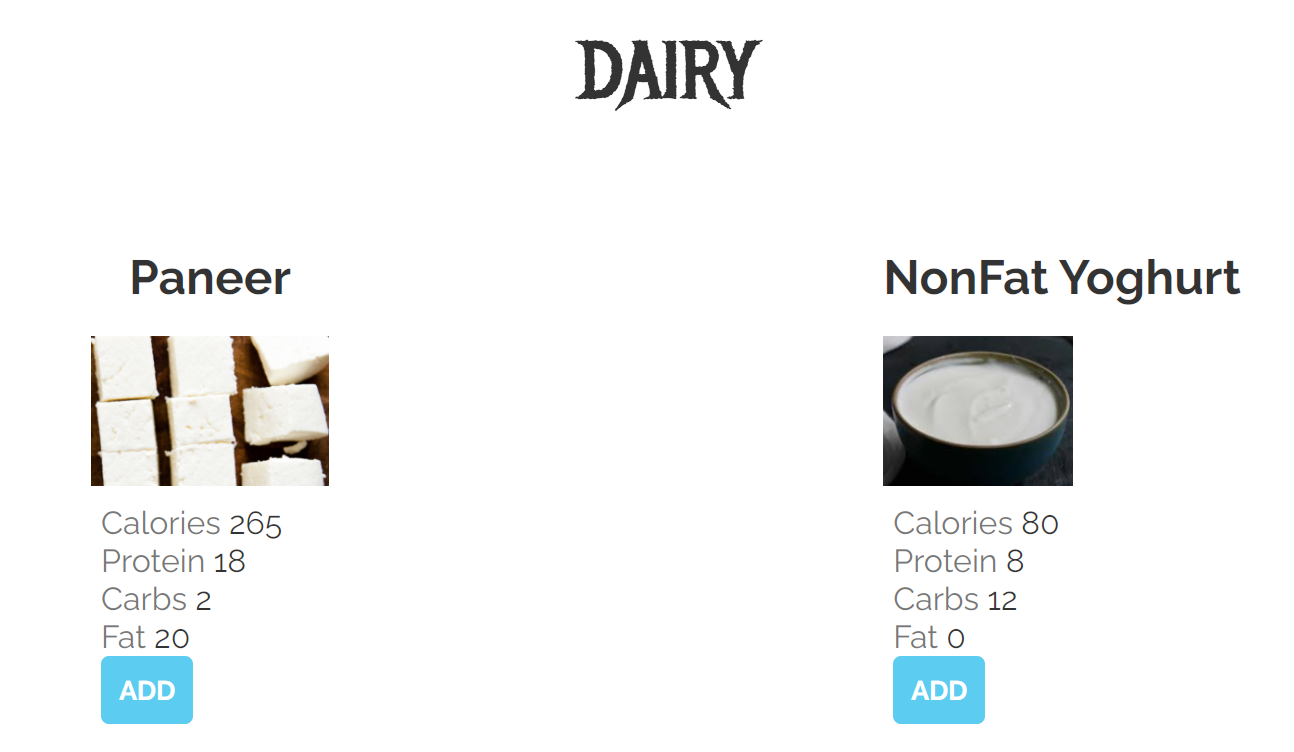}}
\centerline{\includegraphics[width=8cm, height=4cm]{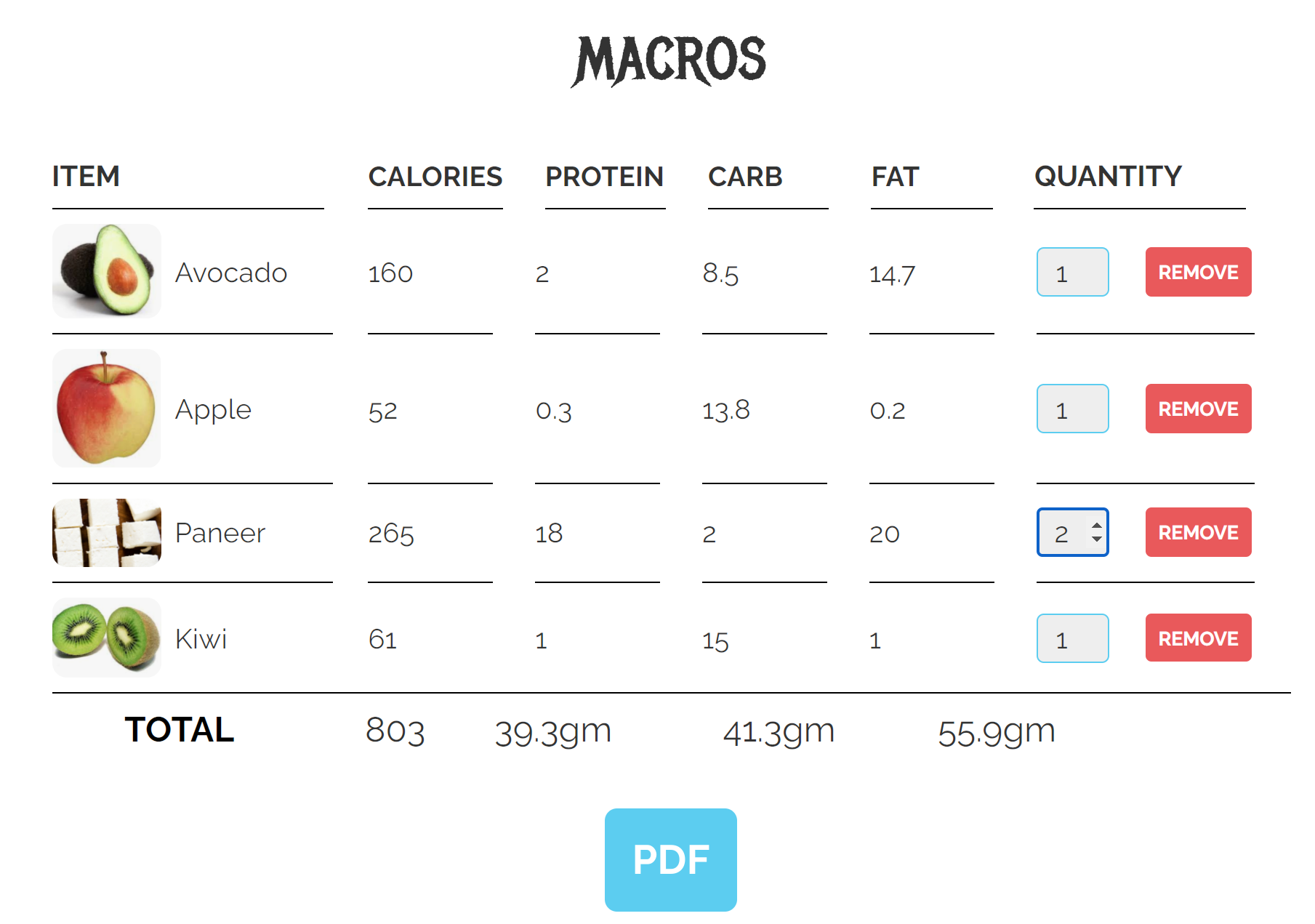}}
\caption{Customizable Diet Plan Genearator}
\end{figure}

\section{Weighing scale}
An IoT-based weighing scale is developed using various components such as HX711 with load cell [32], breadboard, NodeMCU [33], and jumper wire. A flowchart used for building a weighing scale is given in Fig. 10.
\begin{figure}[htbp]
\centerline{\includegraphics[width=8cm, height=5cm]{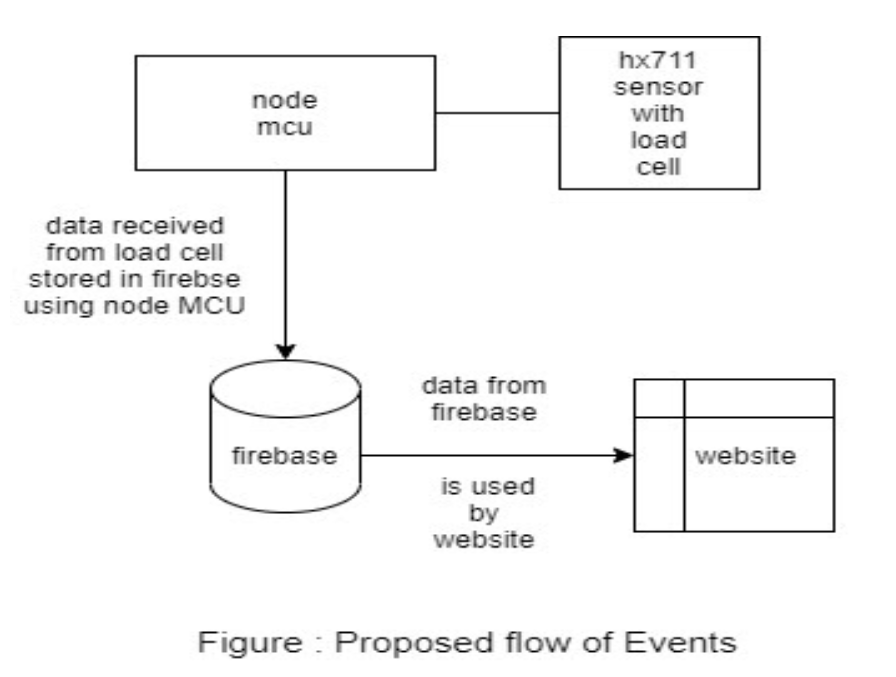}}
\caption{Proposed Flow of events weighing scale}
\end{figure}

The first step is the IoT configuration to extract the weight of an item from the load cell. This can be accomplished by Node MCU controller boards. An item is first placed on the load cell whose Nutrition values the user wants. Then the load cell provides measurable electric signals to the HX711 sensor that further measure the weight of that item. HX711 and load cell then sends this data to the NodeMCU. The second step is the database configuration to store data gathered by sensors. Information can easily put away in the online database using firebase [34] through a program composed using Arduino IDE. The sent data by HX711 and heap cell were transferred and stored by the NodeMCU to Firebase. This information can be utilized for further handling. The third step is to create a web portal for calculating the nutrition values from the weight of an item. The information extraction and playing out specific estimations should be possible with a javascript program.  After doing some calculations on this data the final result can be displayed on the web portal as shown in Figure 11. For example, the user measures 100 grams of oats which contain 363 kcal, 7 grams of fat, 10.3 grams of protein and 60.5 grams of carbohydrates. .
\begin{figure}[htbp]
\centerline{\includegraphics[width=9cm, height=5cm]{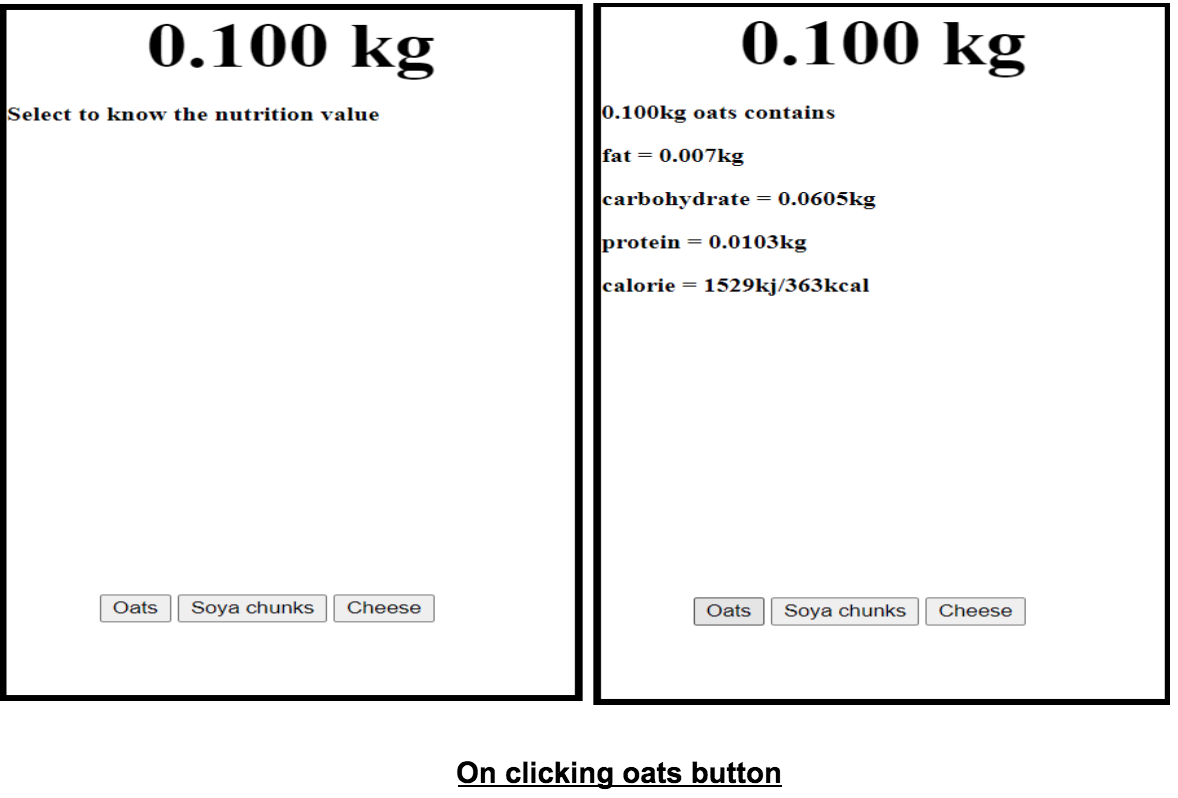}}
\caption{Web Portal Weighing scale}
\end{figure}

The circuit diagram is pictured in Fig. 12. %{application of these pins} 
The various types of pins are connected to perform certain functions.
\begin{figure}[htbp]
\centerline{\includegraphics[width=9cm, height=5cm]{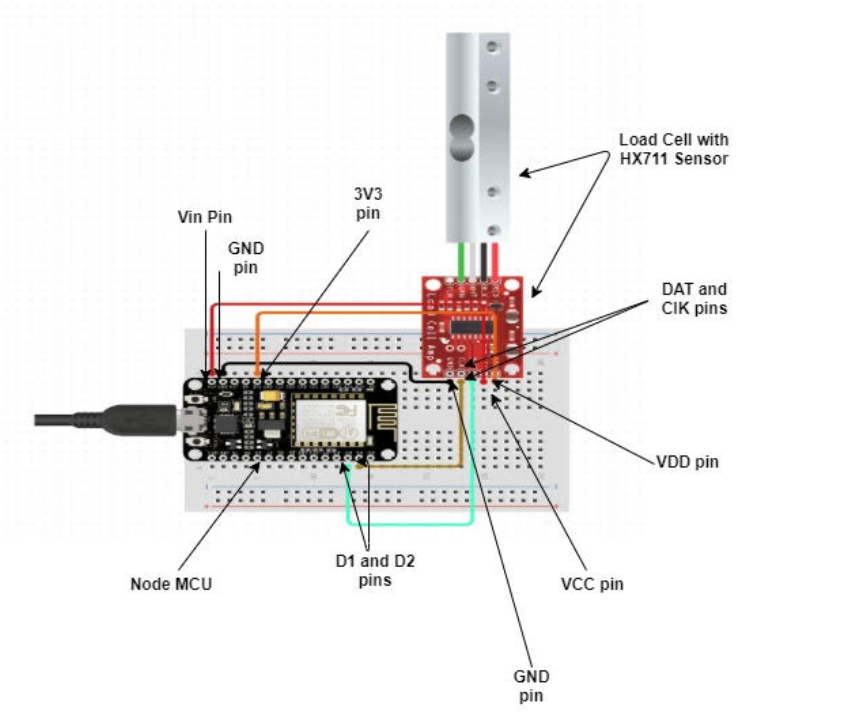}}
\caption{Circuit Diagram}
\end{figure}
VDD is the digital supply voltage used to set the logical level which is connected to the 3V3 pin. The 3V3 pin is the output of the onboard 3.3V regulator for powering components. VCC is the analog voltage to power the load cell. It is connected to the VIN pin which is the voltage input used for the direct power supply. DAT pin and the CLK pin are mainly used for transferring the generated data and connected D2 and D1 pin respectively. It passes on the DAT pin when to go to "low" and communicate information. 
We are primarily willing to look at the values using the pins and codes that are made for the HX711 sensor and align it well. D1 and D2 have specific functions for extracting values from CLK and DAT pins respectively. Gnd pins are ground pins are used to close the electrical circuit and maintain a common reference level through the circuit.
\section{Conclusion}
Obesity resembles a malignancy to the general public and is a significant reason for increasing the risk of various diseases like diabetes, circulatory stress issues and heart diseases. In this research, three machine learning models were created to estimate obesity level, body weight and body fat percentage. Different machine learning algorithms were utilized for preparing the models to give a comparative analysis. Four HPO models are also used for optimization purposes. The results show that Optuna is delivering respectable results in a short amount of time. The selected models were then deployed using a Python Flask. A website framework is also built in with additional highlights, for example, a dashboard and customizable diet plans. An IoT-based weighing scale was also coordinated to quantify the food, which then gives a measure of the calorie and macronutrients of the measured food after preprocessing. In future work, strategies, for example, feature engineering will be utilized to improve the correctness of the models. A significant drawback of this research is that the dataset used is only coming from users living in the South American region. There is a need to prepare separate datasets from different regions and demographic regions to examine the differences. Computer vision applications, for example, muscle to fat ratio and weight forecast using facial or body highlights, are additionally required in the framework which further improves the quality of the MOFit.

\vspace{12pt}

\end{document}